\documentclass[conference,a4paper]{IEEEtran}
\IEEEoverridecommandlockouts
\IEEEpubid{\makebox[\columnwidth]{979-8-3503-2363-4/23/\$31.00~\copyright2023 IEEE\hfill} 
\hspace{\columnsep}\makebox[\columnwidth]{ }}
\usepackage{cite}
\usepackage{amsmath,amssymb,amsfonts}
\usepackage{algorithmic}
\usepackage{graphicx}
\usepackage{textcomp}
\usepackage{xcolor}
\usepackage{listings}
\usepackage{comment}
\usepackage[square,sort&compress,numbers]{natbib}
\newcommand{\footURL}[1]{\footnote{\url{#1}}}
\usepackage{multirow}
\usepackage{hhline}
\usepackage{makecell}
\usepackage{tabularx}
\newcolumntype{Y}{>{\centering\arraybackslash}X}
\newcolumntype{Z}{>{\raggedleft\arraybackslash}X}

\def\BibTeX{{\rm B\kern-.05em{\sc i\kern-.025em b}\kern-.08em
    T\kern-.1667em\lower.7ex\hbox{E}\kern-.125emX}}
\usepackage{hyperref} %[backref=page] % Removed to see if it breaks arxiv :-(
\hypersetup{
colorlinks   = true,
citecolor    = blue,
urlcolor    =  blue
}

\begin{document}

\title{Sinhala-English Parallel Word Dictionary Dataset}

% \author{\IEEEauthorblockN{1\textsuperscript{st} Kasun Wickramasinghe}
\author{\IEEEauthorblockN{Kasun Wickramasinghe, Nisansa de Silva}
\IEEEauthorblockA{\textit{Department of Computer Science \& Engineering} \\
\textit{University of Moratuwa}\\
Moratuwa, Sri Lanka \\
\texttt{\{kasunw.22,NisansaDdS\}@cse.mrt.ac.lk}}
%\and
% \IEEEauthorblockN{2\textsuperscript{nd} Nisansa de Silva}
%\IEEEauthorblockN{Nisansa de Silva}
%\IEEEauthorblockA{\textit{Department of Computer Science and Engineering} \\
%\textit{University of Moratuwa}\\
%Moratuwa, Sri Lanka \\
%nisansadds@cse.mrt.ac.lk}
}

% \author{Anonymous Authors}

\maketitle
\IEEEpubidadjcol
\begin{abstract}
Parallel datasets are vital for performing and evaluating any kind of multilingual task. However, in the cases where one of the considered language pairs is a low-resource language, the existing top-down parallel data such as corpora are lacking in both tally and quality due to the dearth of human annotation. Therefore, for low-resource languages, it is more feasible to move in the bottom-up direction where finer granular pairs such as \emph{dictionary datasets} are developed first. They may then be used for mid-level tasks such as supervised multilingual word embedding alignment. These in turn can later guide higher-level tasks in the order of aligning \emph{sentence} or \emph{paragraph} text corpora used for Machine Translation (MT). Even though more approachable than generating and aligning a massive corpus for a low-resource language, for the same reason of apathy from larger research entities, even these finer granular data sets are lacking for some low-resource languages. We have observed that there is no free and open dictionary data set for the low-resource language, Sinhala. Thus, in this work, we introduce \emph{three} parallel English-Sinhala \emph{word dictionaries} (\verb|En-Si-dict-large|, \verb|En-Si-dict-filtered|, \verb|En-Si-dict-FastText|) which help in multilingual Natural Language Processing (NLP) tasks related to English and Sinhala languages. In this paper, we explain the dataset creation pipeline as well as the experimental results of the tests we have carried out to verify the quality of the data sets. The data sets and the related scripts are available at \url{https://github.com/kasunw22/sinhala-para-dict}.
% [link will be added after blind review].
\end{abstract}

\begin{IEEEkeywords}
parallel corpus, alignment, English-Sinhala dictionary, word embedding alignment, lexicon induction 
\end{IEEEkeywords}

\section{Introduction}

One of the biggest challenges in today's NLP is that most of the contributions are focused on a few high-resource languages~\cite{magueresse2020low, hedderich2020survey,ranathunga2022some,de2019survey} and therefore less priority has been assigned to all the other so-called~\emph{low-resource} languages still with billions of speakers. One of the major reasons for the poor progress made in the improvement of low-resource NLP is the lack of resources on such languages: mainly the lack of experts~\cite{ranathunga2022some,de2019survey} and lack of data~\cite{magueresse2020low,ranathunga2022some,de2019survey}. 
Multilingual tasks associated with low-resource languages lack further for want of enough resources; mainly parallel aligned datasets. Multilingual tasks try to find alignments between the languages by means of the parallel corpora~\cite{mikolov2013exploiting, xing2015normalized, joulin2018loss, lample2019cross, conneau2019unsupervised, feng2020language, fernando2022exploiting}. Alignment tasks can be divided into three main groups according to \citet{magueresse2020low} as word-level, sentence-level, and document-level alignment. 

When it comes to word-level alignment tasks, \emph{lexicon induction} tasks such as inverse consultation (IC)~\cite{saralegi2011analyzing}, lexicon and synonym dictionary developement~\cite{nasution2016constraint} and dictionary induction~\cite{wushouer2015constraint} and, \emph{word embedding alignment}~\cite{mikolov2013exploiting, xing2015normalized, joulin2018loss} are the major tasks that we come across. Machine translation (MT) and multilingual document and sentence alignment tasks (statistical~\cite{brown1990statistical, dara2016yoda, gomes2016first, buck2016quick} and neural-based~\cite{bahdanau2014neural, wu2016google,thillainathan2021fine,fernando2022data}) are two major tasks related to sentence and document-level alignment. Yet another sentence and document-level alignment task that produced impactful results in this domain during the past few years is the introduction of multilingual language models such as XLM~\cite{lample2019cross}, XLMR~\cite{conneau2019unsupervised}, and LaBSE~\cite{feng2020language}.

The objective of this paper is to create and publish multiple \emph{English-Sinhala} parallel dictionaries which can be used for word-level NLP tasks such as lexicon induction and supervised word embedding alignment. Sinhala, being a low-resource language~\cite{ranathunga2022some}, is currently lagging behind in the matter of the very existence of such resources~\cite{de2019survey} and our effort by this is to fill those gaps.

\section{Related Work}
Parallel datasets of good quality are essential for multilingual tasks~\cite{kreutzer2022quality}. The most common type of parallelism we come across is \emph{sentence-level} or \emph{paragraph-level} parallelism where each data point of the dataset is a pair (or group) of sentences or paragraphs. This type of parallelism is useful in tasks such as MT. The other type of parallelism we come across is \emph{word-level} or \emph{token-level} or \emph{finer granular} parallelism where a data point represents a pair (or a group) of words (or tokens) of each language. This type of parallel data sets can be considered as \emph{dictionary datasets} which can be useful in low-level NLP tasks such as supervised multilingual word embedding alignment. By nature, such dictionary datasets are less common compared to the first type of parallel datasets.  Especially when it comes to \emph{low-resource} languages, where even monolingual basic resources such as WordNets are scarce~\cite{wijesiri2014building}, it is quite difficult to find such dictionary data sets which are free and publicly available~\cite{ranathunga2022some}.

We can find several parallel corpora that are related to the Sinhala language. One is the \emph{FLORES} dataset by~\citet{guzman2019flores} which introduces two parallel corpora, one is Nepali-English and the other is Sinhala-English that contains sentences specifically created for MT. \citet{costa2022no} also provides MT-related resources for Sinhala. The work by~\citet{hameed2016automatic} is a Sinhala-Tamil parallel corpus that also consists of sentences. The work by~\citet{banon2020paracrawl}, as well as the work by~\citet{vasantharajan2021tamizhi}, are also sentence corpora that contain Sinhala language resources. Yet, we do not find any \emph{pure} word-level parallel data sets at the moment for the Sinhala language. There is one open-source dictionary \emph{Subasa Ingiya}\footURL{https://subasa.lk/?page_id=3738}~\cite{wasala2008ensitip} but it is a small dictionary that contains about 36000 pairs and contains not only word pairs but also phrases, therefore it is not a \emph{pure} word dictionary. Our effort in this paper is to fill that blank in this research area.

\section{Data Set}

\subsection{Data Set Overview}
\label{dataset_overview}
In this work, we introduce three English-Sinhala parallel dictionaries which will help in word and token-level Sinhala-English multilingual NLP tasks. The unique properties of the dictionary in contrast to sentence and paragraph-level parallel corpora are:

\begin{itemize}
    \item Contains only single words in both languages per pair. This bypasses a number of issues raised by the different levels of inflection between the two languages.
    \item Can contain multiple entries of the same word in both source and target columns (since the same word can have multiple correct translations due to polysemy or otherwise.)
\end{itemize}

We used FastText~\cite{bojanowski2017enriching,joulin2016bag} English\footURL{https://dl.fbaipublicfiles.com/fasttext/vectors-crawl/cc.en.300.vec.gz} and Sinhala\footURL{https://dl.fbaipublicfiles.com/fasttext/vectors-crawl/cc.si.300.vec.gz} monolingual embedding models to extract the dictionary words.
We have experimented with two versions of the datasets.
\begin{itemize}
    \item V1 - Created using only the Sinhala Fasttext vocabulary
    \item V2 - Created using both the Sinhala and English Fasttext vocabularies
\end{itemize}

\subsection{Creation Process}
\label{dict_process}

We are explaining the creation process of the V1 data sets in this section. The procedure for V2 is functionally identical and thus is not explicitly discussed.
First, the word columns of the FastText models are extracted. Then, the English translations for all the Sinhala\footnote{The V1 datasets have been created using only the Sinhala FastText vocabulary, while the V2 datasets used both Sinhala and English vocabularies} words are extracted using \emph{Google Translate API}\footURL{https://cloud.google.com/translate} and all the entries with multi-word English words are removed to ensure the properties mentioned under Section~\ref{dataset_overview}. This is an important filtering step given the difference in the level of inflection between Sinhala and English~\cite{de2019survey}. Here we are providing three versions of dictionaries.

% En-Si-dict-large, En-Si-dict-filtered, En-Si-dict-fasttext
\subsubsection{Dictionary 1: En-Si-dict-large}
\label{dict_1}
\begin{figure}[!htb]
  \centering
  \includegraphics[width=\linewidth]{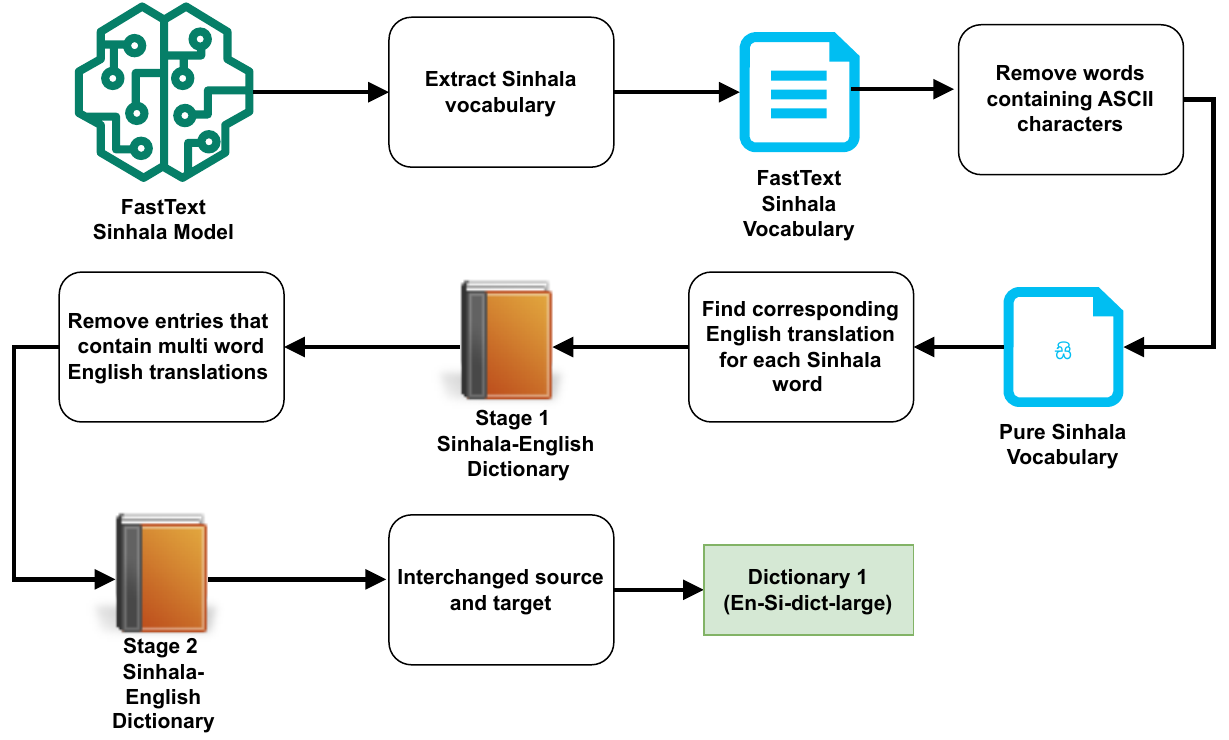}
  \caption{Dictionary 1 Creation Pipeline (For V1 of the Datasets)}
  % \Description{The flow description here}
\label{fig_flow1}
\end{figure}

The first dataset we introduce is an English-Sinhala word dictionary that contains $546,156$ word pairs. This is obtained after removing the multi-word entries from the Google translated results. The relevant pipeline is shown in Figure~\ref{fig_flow1}. 

\paragraph{Extract Sinhala vocabulary}
First, we extract the word column from the FastText Sinhala model. The purpose of this is to make sure the vocabulary in hand would always contain a corresponding vector for the benefit of any future downstream applications. 

\paragraph{Remove words containing ASCII characters}
The words which contain characters other than Sinhala are removed. In this, we removed words (including but not limited to) English and numbers. Owing to the fact that the FastText system is trained on large-scale corpora and the fact that Sinhala writers have historical reasons~\cite{de2019survey} to use code-mixed language, the headwords of ostensibly Sinhala FastText embeddings contain strings that are not of Sinhala. However, when creating a dictionary for Sinhala, we need to remove these to avoid causing harm to downstream applications due to linguistic impurity.

\paragraph{Find corresponding English translation}
Next, the collected Sinhala words are translated into English using the Google Translate API. While there exist translators developed for Sinhala-English pair by a number of previous studies~\cite{thillainathan2021fine,fernando2022data,perera2022improving}, they suffer from weaknesses stemming from lack of generalisability or the limited scope of the training corpora. Further, some also had to be dropped given the fact that there is no publicly available and free API to use. 
Considering all these facts, we opted to use the Google translate API for this study despite the weakness of it pointed out by~\citet{de2019survey} and others.

\paragraph{Remove multi-word entries}
Since the dataset we are creating is a word dictionary, the multi-word English translations are removed. The reason for having multi-word translations for a single Sinhala word at certain points is the fact that Sinhala is a highly inflected language compared to English~\cite{de2015sinhala}. Grammatical cases such as \textit{dative}, \textit{genitive}, and \textit{ablative} which are expressed with multiple words in English are expired with an inflection of the root word in Sinhala~\cite{de2019survey}. Given that we are targeting downstream applications which may not have the ability to process multi-word forms, it was decided to remove them from our solution.  

\paragraph{Interchange source and target}
At this point, we change the first column of the dictionary to English and the second column to Sinhala. This is simply done for the purpose of ease of the subsequent variations that we discuss in Section~\ref{dict_2} and Section~\ref{dict_3}.

\subsubsection{Dictionary 2: En-Si-dict-filtered}
\label{dict_2}
Then a Sinhala word frequency analysis is done using three Sinhala corpora and based on the word frequency information extracted using those corpora, the first dictionary we created in section~\ref{dict_1} is pruned and the second dictionary is built. That process is shown in Figure~\ref{fig_flow2}. When it comes to the corpora used for the Sinhala word frequency analysis, one corpus\footURL{https://osf.io/a5quv/} is by~\citet{upeksha2015implementing,upeksha2015comparison} which was created using web crawling, the second one is a corpus based on \textit{Jathaka Stories}\footURL{https://bit.ly/JathakaTxt} and the third one is based on web crawled news articles\footURL{https://bit.ly/3osodBj}.

\subsubsection{Dictionary 3: En-Si-dict-FastText}
\label{dict_3}
\begin{figure}[!htb]
  \centering
  \includegraphics[width=\linewidth]{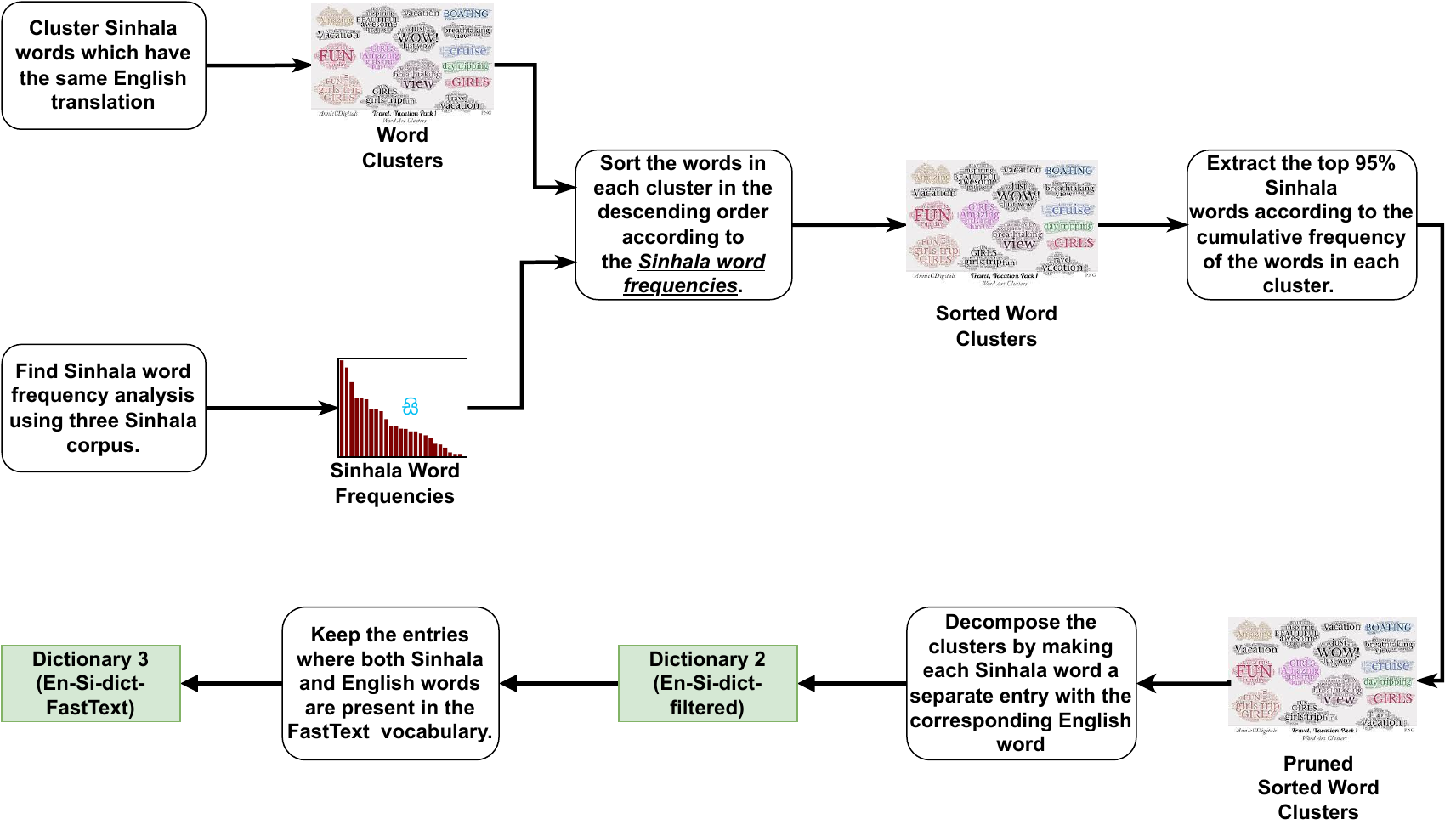}
  \caption{Dictionary 2 and Dictionary 3 Creation Pipeline (For V1 of the Datasets)}
  % \Description{A woman and a girl in white dresses sit in an open car.}
\label{fig_flow2}
\end{figure}

This dictionary contains word pairs that are only present in the FastText word embedding models. All entries that are not present in FastText English and Sinhala models have been removed from the second dictionary (Section~\ref{dict_2}) to form this dictionary. This is the 3rd dictionary we introduce. The creation pipeline is shown in the same Figure~\ref{fig_flow2}

\paragraph{Cluster Sinhala words}
Make clusters of Sinhala words that share the same English word. By doing this we can identify the Sinhala words that go together and it helps with the pruning process we are conducting in a later stage. 

\paragraph{Find Sinhala word frequencies using corpora}
Conduct a Sinhala word frequency analysis using the three Sinhala corpora. The purpose of this is to find the usage frequency of Sinhala words which helps us to do our frequency-based pruning.

\paragraph{Sort each Sinhala word cluster}
By means of the word frequency, each Sinhala word cluster is sorted in the descending order of word frequencies. After this, we have the most frequently used words on the top of each word cluster and infrequent words at the bottom.

\paragraph{Extract top 95\% from each cluster}
From the sorted clusters, the top 95\% most frequent words get selected by cumulative frequency and the rest is pruned. This is to avoid incorrect translation entries. As mentioned by~\citet{de2019survey} and others, there are inconsistencies with Google Translate API, this pruning is to mitigate some of the said translation errors. 

\paragraph{Deconstruct the clusters into a dictionary}
Then we simply convert those pruned cluster results into a dictionary. 

\paragraph{Prune based on the FastText vocabularies}
The dictionary at this point is further pruned based on both the English and the Sinhala word pairs present in the respective FastText vocabularies. This dataset is specifically created for English-Sinhala multilingual downstream tasks related to FastText word vectors.

\section{Statistics}
The statistics of the dataset are shown in Table~\ref{stat_table}. We have shown the unique word percentage with and without stop-words and, the lookup-precision with respect to the FastText vocabularies as described in Equation~\ref{lookup_precision_eqn}. Spacy\footURL{https://spacy.io/models/en} library has been used for stop-word removal wherever necessary.

The \emph{Look-up Precision}, $P_{L}$ means, the proportion of \emph{a word present in the FastText vocabulary, given that word is present in our dictionary}. It is explained in Equation~\ref{lookup_precision_eqn}.

\begin{equation} 
\label{lookup_precision_eqn}
\begin{aligned}
P_{L} = Pr\Big(\frac{\text{\emph{word present in the vocabulary}}}{\text{\emph{word present in the dictionary}}}\Big)
\end{aligned}
\end{equation}

Note that in V1 datasets the \emph{Look-up Precision} ($P_{L}$) of Sinhala is 100\% because only the Sinhala FastText vocabulary is used for the dataset creation process there. 

\begin{table}[!htb]
  \caption{Dataset Statistics}
  \label{stat_table}
  \begin{tabularx}{0.49\textwidth}{llZZZZZ}
    % \toprule
    \hline
    \multirow{2}{*}{Dictionary} & \multirow{2}{*}{Language} & \multicolumn{2}{c}{Entries} & \multicolumn{2}{c}{\makecell{Unique\% \\ w.r.t. stopwords}}  & \multirow{2}{*}{$P_{L}$\%}\\
    \hhline{~~----~}
     &  & Unique  & Total  & With  & Without  & \\
    \hline
    \multirow{1}{6em}{En-Si-dict-large-V1} & English & 134771 & 546156 & 24.7 & 26.4 & 54.1\\ 
    & Sinhala & 546144 & 546156 & 99.9 & 99.9 & 100.0\\
    
    \hline
    \multirow{1}{6em}{En-Si-dict-filtered-V1} & English & 90988 & 195255 & 46.6 & 47.8 & 44.7\\
    & Sinhala & 195247 & 195255 & 99.9 & 99.9 & 100.0\\

    \hline
    \multirow{1}{6em}{En-Si-dict-FastText-V1} & English & 41080 & 136898 & 30.0 & 31.0 & 100.0\\
    & Sinhala & 136896 & 136898 & 99.9 & 99.9 & 100.0\\

    \hline
    \hline

    \hline
    \multirow{1}{6em}{En-Si-dict-large-V2} & English & 915058 & 1368416 & 66.9 & 68.8 & 78.7\\
    & Sinhala & 1030443 & 1368416 & 75.3 & - & 53.0\\
    
    \hline
    \multirow{1}{6em}{En-Si-dict-filtered-V2} & English & 271298 & 332943 & 81.5 & 81.9 & 58.7\\
    & Sinhala & 159405 & 332943 & 47.9 & - & 90.3\\

    \hline
    \multirow{1}{6em}{En-Si-dict-FastText-V2} & English & 159361 & 213463 & 74.7 & 75.2 & 100.0\\
    & Sinhala & 88578 & 213463 & 41.5 & - & 100.0\\
    
    % \bottomrule
    \hline
  \end{tabularx}

    \begin{itemize}
        \item V1 - datasets created by translating the Sinhala Fasttext vocabulary to English
        \item V2 - datasets created by using both Sinhala and English Fasttext vocabularies
    \end{itemize}
\end{table}

\section{Experiments} 
\label{experiments}

To experiment with the quality of our dictionary, we have done a \emph{look-up precision} (token availability) check using En-Si parallel corpora, \verb|WikiMedia|~\cite{tiedemann2012parallel} and \verb|TED2020|~\cite{reimers-2020-multilingual-sentence-bert} as the baseline data sets; which are assumed to be \emph{parallel}, \emph{aligned} and having exact translations rather than having \emph{paraphrased} translations (translations of a paraphrased version of the original source sentences).
The \verb|WikiMedia V20210402|\footURL{https://object.pouta.csc.fi/OPUS-wikimedia/v20210402/tmx/en-si.tmx.gz} data set contains 7.9k aligned parallel sentences in English and Sinhala that are Wikipedia translations published by the Wikimedia Foundation and their article translation 
system\footURL{https://opus.nlpl.eu/wikimedia-v20210402.php}. The  \verb|TED2020 V1|\footURL{https://object.pouta.csc.fi/OPUS-TED2020/v1/tmx/en-si.tmx.gz} data set contains 1k sentences which is a crawl of nearly 4000 \textit{TED} and \textit{TED-X} transcripts from July 2020. The transcripts have been translated by a global community of volunteers\footURL{https://opus.nlpl.eu/TED2020-v1.php}.

%\subsection{Quantitative Evaluation} 
%\label{evaluation}

\begin{figure}[!htb]
  \centering
  \fbox{\includegraphics[width=0.96\linewidth]{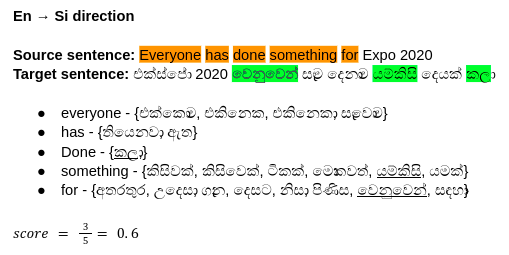}}
  \caption{Example of the Quantitative Evaluation Score - Simple Lookup Score (In the NN-Lookup Scoring method, the target word space is extended with the ten closest words found in the FastText vocabulary of the available target words - So the target word space will contain words that are not available in our dictionary)}
  % \Description{A woman and a girl in white dresses sit in an open car.}
\label{score_example}
\end{figure}

To estimate how good our datasets are, we have defined two scoring criteria (\emph{simple lookup score} and \emph{nearest neighbour lookup score - NN-lookup}) inspired by the widely used \textit{ROUGE-1}~\cite{lin2004rouge}.  The \emph{per-sentence} score we have defined is given in Equation~\ref{score_1} and an illustrated example of the scoring method is shown in Figure~\ref{score_example}.

\begin{equation} 
\label{score_1}
\begin{aligned}
score= \frac{N_{c}}{N_{t}}
\end{aligned}
\end{equation}

Where,
\begin{itemize}
    \item $N_{t}$ - Total number of source sentence words present in the source side of the dictionary
    \item $N_{c}$ - Number of target sentence words present in the word space formed from all the respective target language words (and the 10 nearest neighbours of each target word - for \emph{Nearest Neighbor Lookup score}) of above $N_{t}$ source words.
\end{itemize}

The experiment has been conducted in both directions as English to Sinhala and Sinhala to English. We do not count the sentences where $N_{t}$ is 0. The average score for the entire corpus estimates the quality of our dictionary. Also, the experiment was conducted in three different setups.

\subsubsection{Experimental Setup 1}
\label{exp_setup_1}
Use all the words of the sentences as they are for the experiment (Only case changes have been used).

\subsubsection{Experimental Setup 2}
\label{exp_setup_2}
Remove all the stop-words from the sentences and do the experiment. For the English language, the Spacy\footURL{https://spacy.io/models/en} stop-word list was used and for the Sinhala language, the stop-words list\footURL{https://bit.ly/3wFi0Wf} proposed by~\citet{lakmal2020word} was used.

% \subsubsection{Experimental Setup 3}
% \label{exp_setup_3}
% We remove stopwords only in English sentences and keep all the words in the Sinhala sentences. The reason behind this is that most of the Sinhala stopwords do have a meaning when they are isolated~\cite{wijeratne2020sinhala} but in English, the majority of the stopwords do not.

\section{Results and Discussion}

The results of the experiments done on the three dictionaries we created (see Section~\ref{dict_process}) with the three different setups that are explained under Section~\ref{experiments} have been reported in Table~\ref{exp_results_wikimedia} and Table~\ref{exp_results_ted}. For each dataset, with each setup, the average \emph{lookup-precision score} (Equation~\ref{score_1}) has been calculated in both directions (i.e. English to Sinhala and Sinhala to English).

\begin{table}[!htb]
  \caption{Experimental Results on Wikimedia dataset}
  \label{exp_results_wikimedia}
  % \begin{tabular}{ccccc}
  \begin{tabular}{cp{2em}p{2em}p{2em}p{3em}p{3em}}
    \hline
    % \toprule
    Dictionary & Setup & Source & Target & Average simple-lookup Score & Average NN-lookup Score\\
    \hline
    \multirow{4}{*}{\makecell{Dataset 1-V1\\(En-Si-dict-large-V1)}} & \multirow{2}{4em}{Setup 1} & En & Si & 0.2552 & 0.4175\\
    & & Si & En & 0.3360 & 0.4764 \\
    \hhline{~-----}
    & \multirow{2}{4em}{Setup 2} & En & Si & 0.2552 & 0.3694 \\
    & & Si & En & 0.3267 & 0.4660 \\
    % & \multirow{2}{4em}{Setup 3} & En & Si & 0.2552 & \\
    % & & Si & En & 0.3360 & \\
    
    \hline
    \multirow{4}{*}{\makecell{Dataset 2-V1\\(En-Si-dict-filtered-V1)}} & \multirow{2}{4em}{Setup 1} & En & Si & 0.3340 & 0.4546 \\
    & & Si & En & 0.4086 & 0.5053 \\
    \hhline{~-----}
    & \multirow{2}{4em}{Setup 2} & En & Si & 0.3417 & 0.4147 \\
    & & Si & En & 0.3984 & 0.4915 \\
    % & \multirow{2}{4em}{Setup 3} & En & Si & 0.3417 & \\
    % & & Si & En & 0.4086 & \\
    
    \hline
    \multirow{4}{*}{\makecell{Dataset 3-V1\\(En-Si-dict-FastText-V1)}} & \multirow{2}{4em}{Setup 1} & En & Si & 0.3328 & 0.4535 \\
    & & Si & En & 0.4088 & 0.5064 \\
    \hhline{~-----}
    & \multirow{2}{4em}{Setup 2} & En & Si & 0.3406 & 0.4136 \\
    & & Si & En & 0.3983 & 0.4932 \\
    % & \multirow{2}{4em}{Setup 3} & En & Si & 0.3406 & \\
    % & & Si & En & 0.4088 & \\

    \hline
    \hline
    % \hline[3pt]
        
    % \hline
    \multirow{4}{*}{\makecell{Dataset 1-V2\\(En-Si-dict-large-V2)}} & \multirow{2}{4em}{Setup 1} & En & Si & 0.3666 & \textbf{0.5056} \\
    & & Si & En & 0.4220 & 0.5356 \\
    \hhline{~-----}
    & \multirow{2}{4em}{Setup 2} & En & Si & 0.3772 & 0.4606 \\
    & & Si & En & 0.4068 & 0.5207 \\
    % & \multirow{2}{4em}{Setup 3} & En & Si & 0.3772 & 0.46058 \\
    % & & Si & En & 0.4220 & 0.53564 \\
    
    \hline
    \multirow{4}{*}{\makecell{Dataset 2-V2\\(En-Si-dict-filtered-V2)}} & \multirow{2}{4em}{Setup 1} & En & Si & 0.3781 & 0.4988\\
    & & Si & En & 0.4809 & 0.5825\\
    \hhline{~-----}
    & \multirow{2}{4em}{Setup 2} & En & Si & \textbf{0.3854} & 0.4458\\
    & & Si & En & 0.4620 & 0.5647\\
    % & \multirow{2}{4em}{Setup 3} & En & Si & 0.3854 & 0.4458\\
    % & & Si & En & 0.4809 & 0.5825\\
    
    \hline
    \multirow{4}{*}{\makecell{Dataset 3-V2\\(En-Si-dict-FastText-V2)}} & \multirow{2}{4em}{Setup 1} & En & Si & 0.3766 & 0.4443\\
    & & Si & En & \textbf{0.4810} & 0.5658\\
    \hhline{~-----}
    & \multirow{2}{4em}{Setup 2} & En & Si & 0.3838 & 0.4983\\
    & & Si & En & 0.4617 & \textbf{0.5838}\\
    % & \multirow{2}{4em}{Setup 3} & En & Si & 0.3838 & 0.4983\\
    % & & Si & En & 0.4810 & 0.5658\\
    
    % \bottomrule
    \hline
  \end{tabular}
\end{table}

% \multirow{1}{6em}{En-Si-dict-large-V1}

\begin{table}[!htb]
  \caption{Experimental Results on TED2020 dataset}
  \label{exp_results_ted}
  % \begin{tabular}{cccccc}
  \begin{tabular}{cp{2em}p{2em}p{2em}p{3em}p{3em}}
    % \toprule
    \hline
    Dictionary & Setup & Source & Target & Average simple-lookup Score & Average NN-lookup Score\\
    \hline
    \multirow{4}{*}{\makecell{Dataset 1-V1\\(En-Si-dict-large-V1)}} & 
    \multirow{2}{4em}{Setup 1} & En & Si & 0.2253 & 0.4052 \\
    & & Si & En & 0.2950 & 0.4688 \\
    \hhline{~-----}
    & \multirow{2}{4em}{Setup 2} & En & Si & 0.2648 & 0.4009 \\
    & & Si & En & 0.2828  & 0.4601\\
    % & \multirow{2}{4em}{Setup 3} & En & Si & 0.2648 &\\
    % & & Si & En & 0.2950 &\\
    
    \hline
    \multirow{4}{*}{\makecell{Dataset 2-V1\\(En-Si-dict-filtered-V1)}} &
    \multirow{2}{4em}{Setup 1} & En & Si & 0.2900 & 0.4275\\
    & & Si & En & 0.3640 & 0.4947\\
    \hhline{~-----}
    & \multirow{2}{4em}{Setup 2} & En & Si & 0.3385 & 0.4242\\
    & & Si & En & 0.3501 & 0.4869\\
    % & \multirow{2}{4em}{Setup 3} & En & Si & 0.3385 &\\
    % & & Si & En & 0.3640 &\\
    
    \hline
    \multirow{4}{*}{\makecell{Dataset 3-V1\\(En-Si-dict-FastText-V1)}} &
    \multirow{2}{4em}{Setup 1} & En & Si & 0.2900 & 0.4275\\
    & & Si & En & 0.3662 & 0.4980\\
    \hhline{~-----}
    & \multirow{2}{4em}{Setup 2} & En & Si & 0.3385 & 0.4246\\
    & & Si & En & 0.3524 & 0.4904\\
    % & \multirow{2}{4em}{Setup 3} & En & Si & 0.3385 &\\
    % & & Si & En & 0.3662 &\\

    \hline
    \hline

    \multirow{4}{*}{\makecell{Dataset 1-V2\\(En-Si-dict-large-V1)}} &
    \multirow{2}{4em}{Setup 1} & En & Si & 0.3296 &0.5050 \\
    & & Si & En & 0.4003 & 0.5514\\
    \hhline{~-----}
    & \multirow{2}{4em}{Setup 2} & En & Si & \textbf{0.3859} & \textbf{0.5121}\\
    & & Si & En & 0.3874 & 0.5403\\
    % & \multirow{2}{4em}{Setup 3} & En & Si & 0.3859 & 0.5121\\
    % & & Si & En & 0.4003 & 0.5514\\
    
    \hline
    \multirow{4}{*}{\makecell{Dataset 2-V2\\(En-Si-dict-filtered-V1)}} &
    \multirow{2}{4em}{Setup 1} & En & Si & 0.3269 & 0.4699\\
    & & Si & En & 0.4329 & 0.5498 \\
    \hhline{~-----}
    & \multirow{2}{4em}{Setup 2} & En & Si & 0.3804 & 0.4713\\
    & & Si & En & 0.4190 & \textbf{0.5585}\\
    % & \multirow{2}{4em}{Setup 3} & En & Si & 0.3804 & 0.4713\\
    % & & Si & En & 0.4329 & 0.5498\\
    
    \hline
    \multirow{4}{*}{\makecell{Dataset 3-V2\\(En-Si-dict-FastText-V1)}} &
    \multirow{2}{4em}{Setup 1} & En & Si & 0.3272 & 0.4706 \\
    & & Si & En & \textbf{0.4368} & 0.5556 \\
    \hhline{~-----}
    & \multirow{2}{4em}{Setup 2} & En & Si & 0.3810 & 0.4718 \\
    & & Si & En & 0.4231 & 0.5638 \\
    % & \multirow{2}{4em}{Setup 3} & En & Si & 0.3810 & 0.4718\\
    % & & Si & En & 0.4368 & 0.5556 \\
    \hline
    % \bottomrule
  \end{tabular}
\end{table}

%\section{Discussion and Conclusion}
The V1 datasets have been created using only the Sinhala FastText vocabulary and therefore the lookup precision of the Sinhala language is 100\% in these datasets (Table~\ref{stat_table}). For the same reason, the unique Sinhala word percentage is significantly high in V1 datasets. This means, that for a given Sinhala word, most of the time, there is only one English translation in the dictionary which indicates the dictionaries are missing a great portion of synonymous meanings of Sinhala words. This in turn leads to a low \emph{look-up precision} score (Section~\ref{experiments}) in Sinhala to English direction. Therefore, the lower the unique word percentage, the higher the score we can expect. It can be clearly seen since the best scores have been shown in Table~\ref{exp_results_wikimedia} and Table~\ref{exp_results_ted} by the entries that have the lowest unique word percentages in Table~\ref{stat_table}.

Apart from that, the metric we used to measure the quality of our datasets does not give extremely high scores. One reason for this observation is the fact that the two languages are syntactically and semantically quite different and therefore the parallel sentence pairs may not fully tally with each other. Another reason we can postulate is that this is due to not having a low enough unique word percentage as we discussed earlier in this section.

\section{Future Work}

We have identified the following as our next steps to improve the content and the quality of our data sets.

\begin{itemize}
    \item Find more monolingual vocabularies and extend the dataset following the same procedures.
    \item Look for (or derive) better matrices to evaluate a dictionary-type dataset.
    \item Look for techniques to bring the unique word percentage further down (include more alternative meanings to a given source word by explicitly handling polysemy).
\end{itemize}

\section{Conclusion}

One of the challenges we come across in low-resource language NLP tasks~\cite{ranathunga2022some} is the lack of free and publicly available resources such as data sets~\cite{de2019survey}. Even the available resources suffer from quality issues~\cite{kreutzer2022quality}. Sinhala, as such a low-resource language, does have some publicly available parallel corpora~\cite{guzman2019flores, costa2022no, hameed2016automatic, banon2020paracrawl, vasantharajan2021tamizhi} that suit high-level (i.e. sentence and paragraph level) NLP tasks such as MT but it is suffering from not having publicly available finer granular pair data sets. Here we have introduced three English-Sinhala dictionary data sets that only consist of parallel words. We are confident that this will give some energy to the Sinhala language-related lower-level multilingual NLP tasks. We are continuing this research further and are hoping to release another version of these data sets by rectifying the obstacles identified and discussed in this work earlier.

% References
\bibliographystyle{IEEEtranN}
\bibliography{references}

% Nisansa stopped reading here

%\vspace{12pt}

\end{document}